\documentclass[runningheads]{llncs}
\usepackage[T1]{fontenc}
\usepackage{graphicx}
\usepackage{wrapfig}
\usepackage{hyperref}
\usepackage{multirow}
\usepackage{comment}

\begin{document}
\title{Towards A Human-in-the-Loop LLM Approach to Collaborative Discourse Analysis}
\titlerunning{Towards A HITL LLM Approach to Collaborative Discourse Analysis}
%

\author{Clayton Cohn\inst{1}\orcidID{0000-0003-0856-9587} \and
Caitlin Snyder\inst{1}\orcidID{0000-0002-3341-0490} 
\and
Justin Montenegro\inst{2} 
\and
Gautam Biswas\inst{1}\orcidID{0000-0002-2752-3878}}

\institute{Vanderbilt University, Nashville, TN 37240, USA \\
\email{clayton.a.cohn@vanderbilt.edu} \and
Martin Luther King, Jr. Academic Magnet High School, Nashville, TN 37203, USA \\ }

\authorrunning{C. Cohn et al.}
%

\maketitle              

\begin{abstract}

LLMs have demonstrated proficiency in contextualizing their outputs using human input, often matching or beating human-level performance on a variety of tasks. However, LLMs have not yet been used to characterize synergistic learning in students' collaborative discourse. In this exploratory work, we take a first step towards adopting a human-in-the-loop prompt engineering approach with GPT-4-Turbo to summarize and categorize students' synergistic learning during collaborative discourse. Our preliminary findings suggest GPT-4-Turbo may be able to characterize students' synergistic learning in a manner comparable to humans and that our approach warrants further investigation. 

\keywords{LLM \and Collaborative Learning \and Human-in-the-Loop \and Discourse Analysis \and K12 STEM.}
\end{abstract}

\section{Introduction} \label{sec:introduction}

Computational modeling of scientific processes has been shown to effectively foster students' Science, Technology, Engineering, Mathematics, and Computing (STEM+C) learning \cite{hutchins2020c2stem}, but task success necessitates \textit{synergistic learning} (i.e., the simultaneous development and application of science and computing knowledge to address modeling tasks), which can lead to student difficulties \cite{basu2016identifying}. Research has shown that problem-solving environments promoting synergistic learning in domains such as physics and computing often facilitate a better understanding of physics and computing concepts and practices when compared to students taught via a traditional curriculum \cite{hutchins2020c2stem}. Analyzing students' collaborative discourse offers valuable insights into their application of both domains' concepts as they construct computational models \cite{snyder2019analyzing}. 
Unfortunately, manually 
analyzing students' discourse to identify their synergistic processes is time-consuming, and programmatic approaches are needed. 


In this paper, we take an exploratory first step towards adopting a \textit{human-in-the-loop} LLM approach from previous work called \textit{Chain-of-Thought Prompting + Active Learning} \cite{cohn2024chain} (detailed in Section \ref{sec:methods}) to characterize the synergistic content in students' collaborative discourse. We use a large language model (LLM) to summarize conversation segments in terms of how physics and computing concepts are interwoven to support students' model building and debugging tasks
. We evaluate our approach by comparing the LLM's summaries to human-produced ones (using an expert human evaluator to rank them) and by qualitatively analyzing the summaries to discern the LLM's strengths and weaknesses alongside a physics and computer science teacher (the Educator) with experience teaching the C2STEM curriculum (see Section \ref{subsec:curriculum_and_data}). Within this framework, we analyze data from high school students working in pairs to build kinematics models and answer the following research questions: \textbf{RQ1}) How does the quality of human- and LLM-generated summaries and synergistic learning characterizations of collaborative student discourse compare?, and \textbf{RQ2}) What are the LLM's strengths, and where does it struggle, in summarizing and characterizing synergistic learning in physics and computing?


    

As this work is exploratory, due to the small sample size, we aim not to present generalizable findings but hope that our results will inform subsequent research as we work towards forging a human-AI partnership by providing teachers with actionable, LLM-generated feedback and recommendations to help them guide students in their synergistic learning. 

\section{Background} \label{sec:background}

Roschelle and Teasley \cite{roschelle1995construction} 
define collaboration as \textit{“a coordinated, synchronous activity that is a result of a continuous attempt to construct and maintain a shared conception of a problem.”} This development of a shared conceptual understanding necessitates multi-faceted collaborative discourse across multiple dimensions: social (e.g., navigating the social intricacies of forming a consensus \cite{weinberger2006framework}), cognitive (e.g., the development of context-specific knowledge \cite{snyder2019analyzing}), and metacognitive (e.g., socially shared regulation \cite{hadwin2011self}). Researchers have developed and leveraged frameworks situated within learning theory to classify and analyze collaborative problem solving (CPS) both broadly (i.e., across dimensions \cite{meier2007rating}) and narrowly (i.e., by focusing on one CPS aspect to gain in-depth insight, e.g., argumentative knowledge construction \cite{weinberger2006framework}). In this paper, we focus on one dimension of CPS that is particularly important to the context of STEM+C learning: students' cognitive integration of synergistic domains. 

Leveraging CPS frameworks to classify student discourse has traditionally been done through hand-coding utterances. However, this is time-consuming and laborious, leading researchers to leverage automated classification methods such as rule-based approaches, supervised machine learning methods, and (more recently) LLMs \cite{suraworachet2024predicting}. Utilizing LLMs can help extend previous work on classifying synergistic learning discourse, which has primarily relied on the frequency counts of domain-specific concept codes \cite{snyder2019analyzing,hutchins2020c2stem}. In particular, the use of LLMs can help address the following difficulties encountered while employing traditional methods: (1) concept codes are difficult to identify programmatically, as rule-based approaches like regular expressions (regex) have difficulties with misspellings and homonyms; (2) the presence or absence of concept codes is not analyzed in a conversational context; and (3) the presence of cross-domain concept codes is not necessarily indicative of synergistic learning, as synergistic learning requires students to form connections between concepts in both domains. 

Recent advances in LLM performance capabilities have allowed researchers to find new and creative ways to apply these powerful models to education using \textit{in-context learning} (ICL) \cite{brown2020language} (i.e., providing the LLM with labeled instances during inference) in lieu of traditional training that requires expensive parameter updates. One prominent extension of ICL is \textit{chain-of-thought reasoning} (CoT) \cite{wei2022chain}, which augments the labeled instances with ``reasoning chains'' that explain the rationale behind the correct answer and help guide the LLM towards the correct solution. Recent work has found success in leveraging CoT towards scoring and explaining students' formative assessment responses in the Earth Science domain \cite{cohn2024chain}. In this work, we investigate this approach as a means to summarize and characterize synergistic learning in students' collaborative discourse.

\section{Methods} \label{sec:methods}

\begin{figure}[bp]
    \centering
    \includegraphics[width=0.75\textwidth]{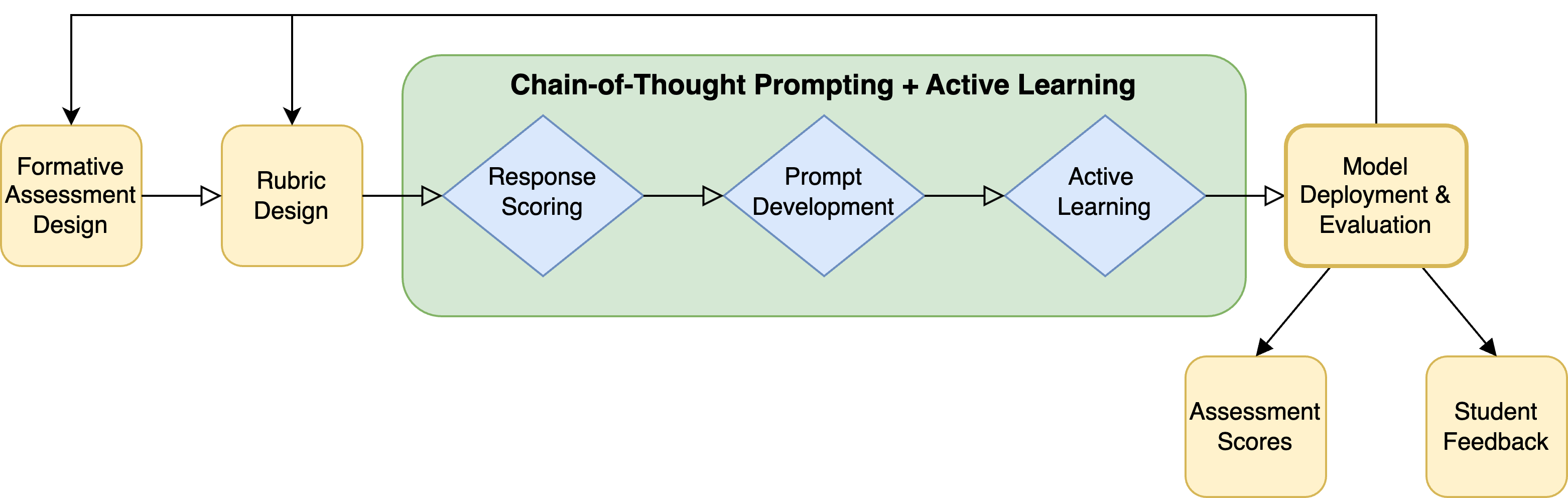}
    \caption{\textit{Chain-of-Thought Prompting + Active Learning}, identified by the green box, where each blue diamond is a step in the Method. Yellow boxes represent the process's application to the classroom detailed in prior work \cite{cohn2024chain}.} 
    \label{fig:approach}
\end{figure}

This paper extends the previous work of 1) Snyder et al. on log-segmented discourse summarization defined by students' model building segments extracted from their activity logs \cite{snyder2024analyzing}, and 2) Cohn et al. on a human-in-the-loop prompt engineering approach called \textit{Chain-of-Thought Prompting + Active Learning} \cite{cohn2024chain} (the Method) for scoring and explaining students' science formative assessment responses. The original Method is a three-step process: 1) \textit{Response Scoring}, where two human reviewers manually label a sample of students' formative assessment responses and identify disagreements (i.e., sticking points) the LLM may similarly struggle with; 2) \textit{Prompt Development}, which employs few-shot CoT prompting to address the sticking points and help align the LLM with the humans' scoring consensus; and 3) \textit{Active Learning}, where a knowledgeable human (e.g., a domain expert, researcher, or instructor) acts as an ``oracle'' and identifies the LLM's reasoning errors on a validation set, then appends additional few-shot instances that the LLM struggled with to the prompt and uses CoT reasoning to help correct the LLM's misconceptions. We illustrate the Method in Figure \ref{fig:approach}. For a complete description of the Method, please see 
\cite{cohn2024chain}. 

In this work, we combine log-based discourse segmentation \cite{snyder2024analyzing} and CoT prompting \cite{cohn2024chain} to generate more contextualized summaries of students' discourse segments to study students' synergistic learning processes by linking their model construction and debugging activities with their conversations during each probl-\\em-solving segment. We provide Supplementary Materials\footnote{\href{https://github.com/oele-isis-vanderbilt/AIED24_LBR}{https://github.com/oele-isis-vanderbilt/AIED24\_LBR}} that include 1) additional information about the learning environment, 2) method application details (including our final prompt and few-shot example selection methodology), 3) a more in depth look at our conversation with the Educator, and 4) a more detailed analysis of the LLM's strengths and weaknesses while applying the Method.


\subsection{STEM+C Learning Environment, Curriculum, and Data} \label{subsec:curriculum_and_data}
Our work in this paper centers on the C2STEM learning environment \cite{hutchins2020c2stem}, where students learn kinematics by building computational models of the 1- and 2-D motion of objects. C2STEM combines block-based programming with domain-specific modeling blocks to support the development and integration of science and computing knowledge as students create partial or complete models that simulate behaviors governed by scientific principles. This paper focuses on the 1-D \textit{Truck Task}, where students use their knowledge of kinematic equations to model the motion of a truck that starts from rest, accelerates to a speed limit, cruises at that speed, then decelerates to come to a stop at a stop sign. 

Our study, approved by our university Institutional Review Board, included 26 consented high school students (aged 14-15) who completed the C2STEM kinematics curriculum. Students' demographic information was not collected as part of this study (we began collecting it in later studies). 
Data collection included logged actions in the C2STEM environment, saved project files, and video and audio data (collected using laptop webcams and OBS software). Our data analysis included 9 dyads (one group had a student who did not consent to data collection, so we did not analyze that group; and we had technical issues with audio data from other groups). The dataset includes 9 hours of discourse transcripts and over 2,000 logged actions collected during one day of the study. Student discourse was transcribed using Otter.ai and edited for accuracy. 

\subsection{Approach} \label{subsec:approach}

We extend the Method, previously used for formative assessment scoring and feedback, to prompt GPT-4-Turbo to summarize segments of students' discourse and identify the \textit{Discourse Category} (defined momentarily) by answering the following question: ``Given a discourse segment, and its environment task context and actions, is the students' conversation best characterized as \textit{physics-focused} (i.e.,
the conversation is primarily focused on the physics domain), \textit{computing-focused} (i.e., the conversation is primarily focused on the computing domain), \textit{physics-and-computing-synergistic} (i.e., students discuss concepts from both domains, interleaving them throughout the conversation, and making connections between them), or \textit{physics-and-computing-separate} (i.e., students discuss both domains but do so separately without interleaving)?'' We use the recently released GPT-4-Turbo LLM (gpt-4-0125-preview) because it provides an extended context window (128,000 tokens). 

We selected 10 training instances and 12 testing instances (10 additional segments were used as a validation set to perform Active Learning) prior to Response Scoring, using stratified sampling to approximate a uniform distribution across Discourse Categories for both the train and test sets. Note that the student discourse was segmented based on which element of the model the students were working on (identified automatically via log data). During Response Scoring, the first two authors of this paper (Reviewers R1 and R2, respectively) independently evaluated the training set segments, classifying each segment as belonging to one of the four Discourse Categories. For each segment the Reviewers disagreed on, the reason for disagreement was noted as a sticking point, and the segment was discussed until a consensus was reached on the specific Discourse Category for that segment. R1 and R2 initially struggled to agree on segments' Discourse Categories (Cohen's $k=0.315$). This is because segments often contained concepts from both domains that may or may not have been interwoven, so it was not always clear which Discourse Category a segment belonged to. Because of this, the Reviewers ultimately opted to label all segments via consensus coding.


During \textit{Prompt Development}, we provided the LLM with explicit task instructions, curricular and environment context, and general guidelines (e.g., instructing the LLM to cite evidence directly from the students' discourse to support its summary decisions and Discourse Category choice). We supplemented the prompt with extensive contextual information not found in previous work \cite{snyder2024analyzing}, including the Discourse Categories, C2STEM variables and their values, physics and computing concepts and their definitions, and students' actions in the learning environment (derived from environment logs). Four labeled instances were initially appended to the prompt as few-shot examples (one per Discourse Category). Active Learning was performed for a total of two rounds over 10 validation set instances, at the end of which one additional few-shot instance was added. 

Before testing, R1 wrote summaries (and labeled Discourse Categories) for the 12 test instances. R2 then compared the human-generated summaries to two LLMs' summaries: GPT-4-Turbo and GPT-4. We compare GPT-4 to GPT-4-Turbo to see which LLM is most promising for use in future work. 
To evaluate RQ1, R2 used ``ranked choice'' to rank the three summaries from best to worst for each test set instance without knowledge of whether the summaries were generated by a human, GPT-4-Turbo, or GPT-4 (the Competitors). Three rankings were used for the scoring: (1) \textit{Wins} (the number of times each Competitor was ranked higher than another Competitor across all instances, i.e., the best Competitor for an individual segment receives two ``wins'' for outranking the other two Competitors for that segment); (2) \textit{Best} (the number of instances each Competitor was selected as the best choice); and (3) \textit{Worst} (the number of instances each Competitor was selected as the worst choice). To answer RQ1, we used the Wilcoxon signed-rank test 
to determine if the difference in rankings between the Human and GPT-4-Turbo's summaries was statistically significant. We also qualitatively compared the differences between the summaries. To answer RQ2, we performed qualitative analysis using the constant comparative method and interviewed the Educator 
to derive GPT-4-Turbo's strengths and weaknesses using the Method for our task. 

\section{Findings} \label{sec:findings}

To answer \textbf{RQ1}, we first present the test results for the \textit{Wins}, \textit{Best}, and \textit{Worst} rankings for the 12 test set instances for all three Competitors in Table \ref{tab:results}. 
For all three metrics, the human outperformed GPT-4-Turbo and GPT-4-Turbo outperformed GPT-4, as evaluated by R2. While ranking, R2 remarked on several occasions that GPT-4-Turbo's responses stood out as being the most detailed and informative regarding the students' problem-solving processes. GPT-4-Turbo also correctly identified a segment's Discourse Category and explained why it did not belong to another category even though the distinction was nuanced. Conversely, GPT-4 suffered from hallucinations in a number of instances and failed to produce a \textit{Best} summary. For example, GPT-4 included a physics concept in its summary that was not part of the discourse and cited irrelevant evidence (e.g., it cited a student who said ``I got it'' as evidence of a computing concept). 
R2 also remarked GPT-4 was prone to generating summaries that lacked depth and detail. There were no discernible trends in the LLMs' abilities to classify segments across different Discourse Categories.

\begin{wraptable}{L}{5cm}
    \begin{tabular}{c|c|c|c}
        & R1 & GPT-4-Turbo & GPT-4 \\
        \hline
        Wins & \textbf{17} & 12 & 7 \\
        Best & \textbf{8} & 4 & 0 \\
        Worst & \textbf{3} & 4 & 5 \\
    \end{tabular}
    \caption{Competitors' rankings across all test set instances. The best-performing Competitor for each segment is in \textbf{bold}.}
    \label{tab:results}
\end{wraptable} 


To quantify our answer for \textbf{RQ1}, we tested the differences between the three Competitors' raw rankings via Wilcoxon signed-rank tests, which yielded $p=0.519$ and $p=0.266$ comparing the Human to GPT-4-Turbo and GPT-4-Turbo to GPT-4, respectively; and $p=0.052$ comparing the Human to GPT-4, implying the ranking differences were not significant at the $p=0.05$ level for all three comparisons. Future work with a larger sample size is necessary to determine if humans outperform GPT-4-Turbo or GPT-4, as our study cannot rule this out (especially given the low p-value comparing the Human to GPT-4). 
Contrary to the quantitative findings, our qualitative analysis revealed that GPT-4-Turbo exhibited several strengths relative to the Human, such as providing greater detail and explaining why a Discourse Category was not appropriate for a given segment. This is especially useful in supporting classroom instructors and generating automated, adaptive scaffolding for students. 
To answer \textbf{RQ1}, the Method enables GPT-4-Turbo to perform 
similarly to humans for this task and dataset but both Competitors exhibit nuances that warrant further investigation.


We answer \textbf{RQ2} by analyzing GPT-4-Turbo's test set generations via the constant comparative method to discern its strengths and weaknesses and by integrating insights from our conversation with the Educator.
The LLM consistently followed prompt instructions, cited relevant discourse pieces similarly to humans (often citing the exact same discourse pieces as R1), adhered to the CoT reasoning chains outlined in the few-shot examples, and selectively extracted relevant information from the segments for summarization. These results corroborate previous findings \cite{cohn2024chain}. Notably, the LLM seamlessly integrated the additional context (see Section \ref{subsec:approach}) 
and exhibited accurate coreference resolution, correctly identifying entities like physics and computing concepts when ambiguous pronouns such as ``it'' or ``that'' were used. 
The LLM effectively recognized ambiguous segments, and explained when multiple Discourse Categories may be applicable and why the less relevant one was not chosen. The model also showcased adept zero-shot identification of concepts defined in the prompt but not used in the reasoning chains
. 

The Educator highlighted the LLM's ability to pinpoint specific items the human may have missed. In one instance, the Educator was shown an LLM-generated segment summary that he initially believed to be \textit{physics-and-comput\-ing-synergistic}, but he later agreed with the LLM that the segment was best categorized as \textit{physics-and-computing-separate}, as the students were merely discussing the domains sequentially and not interleaving and forming connections between the cross-domain concepts. The Educator also valued the LLM's ability to highlight when students may be in need of teacher assistance and provided several ideas for enhancing the human-AI partnership by using the LLM's summaries to generate actionable insight to support students' STEM+C learning (e.g., using the LLM's summaries to create a graphical timeline to capture students' conceptual understanding).

Despite GPT-4-Turbo's capabilities, there are notable areas for improvement. The autoregressive nature of LLMs introduces challenges related to reliance on keywords and phrases. As the LLM considers every token generated previously in subsequent iterations, any hallucinations (or misinterpretations of the prompt or its own generation) can propagate forward and compromise the overall integrity of its response. An instance of this occurred when the LLM fixated on two physics concepts during summarization when the segment was almost entirely focused on the computing domain. GPT-4-Turbo's initial focus on the physics concepts caused it to label the segment as \textit{physics-and-computing-synergistic}, even though both Reviewers and GPT-4 all considered the segment to be \textit{computing-focused}. 
Additionally, the LLM's ability to integrate environment actions in its summaries was limited, often addressing them superficially without connecting them to the broader discourse context. 
The Educator also suggested the LLM should consider the temporality of segments by incorporating ``pause'' identification and duration from prosodic audio via timestamps and suggested highlighting instances where students expressed uncertainty by saying things like ``Um...'' or ``I'm still a little stumped'', as both may help teachers identify students' difficulties. 

\section{Conclusion, Limitations, and Future work} \label{sec:limitations_future_word}

The primary limitation of this exploratory study is its small test sample size (12 segments). Additionally, only one researcher ranked the three sets of test set summaries (one human-generated and two LLM-generated). The constraints we faced were the time-cost of manually labeling, summarizing, analyzing, and evaluating the individual segments and summaries (up to two hours and $\approx$256 tokens per segment summary). While these results cannot be generalized, we have demonstrated the Method's potential for summarizing and characterizing students' synergistic discourse in a manner that can deepen educators' insights into students' cross-domain conceptual understandings. In future work, we will conduct an extensive evaluation to test the generalizability of our approach, including evaluating our Method's performance across various learning tasks.

\begin{credits}

\subsubsection{\ackname} 
This work is supported under National Science Foundation awar\-ds DRL-2112635 and IIS-2327708.

\subsubsection{\discintname}
The authors have no competing interests to declare.
\end{credits}



\bibliographystyle{splncs04}
\bibliography{references}

\end{document}